\documentclass{article}

\usepackage[main, final, nonatbib]{neurips_2026}

\usepackage[utf8]{inputenc} 
\usepackage[T1]{fontenc}    
\usepackage{hyperref}       
\usepackage{url}            
\usepackage{booktabs}       
\usepackage{amsfonts}       
\usepackage{nicefrac}       
\usepackage{microtype}      
\usepackage{xcolor}         
\usepackage{amsmath}

\usepackage{multirow}
\usepackage{graphicx}

\usepackage{amssymb}
\usepackage{textcomp}
\usepackage{gensymb}
\usepackage{wrapfig}
\usepackage{subcaption}
\usepackage{makecell}

\title{Rethinking Vector Field Learning for \\ Generative
Segmentation}

\author{%
  Chaoyang Wang$^{1}$ \quad Yaobo Liang$^{1}$ \quad Boci Peng$^{1}$ \quad Fan Duan$^{2}$ \quad Yunhai Tong$^{1}$ \\
  $^{1}$State Key Lab of General AI, School of Intelligence Science and Technology, Peking University \\
  $^{2}$Tsinghua University \\
  \texttt{cywang@stu.pku.edu.cn} \\
}

\begin{document}

\maketitle

\begin{abstract}

Taming diffusion models for generative segmentation has attracted increasing attention. 
While existing approaches primarily focus on architectural tweaks or training heuristics, there remains a limited understanding of the intrinsic mismatch between continuous flow matching objectives and discrete perception tasks.
In this work, we revisit diffusion segmentation from the perspective of vector field learning.
We identify two key limitations of the commonly used flow matching objective: gradient vanishing and trajectory traversing, which result in slow convergence and poor class separation.
To tackle these issues, we propose a principled vector field reshaping strategy that augments the learned velocity field with a detached distance-aware correction term.
This correction introduces both attractive and repulsive interactions, enhancing gradient magnitudes near centroids while preserving the original diffusion training framework.
Furthermore, we design a computationally efficient, quasi-random centroid encoding scheme inspired by Kronecker sequences, which integrates seamlessly with an end-to-end pixel neural field framework for pixel-level semantic alignment.
Extensive experiments consistently demonstrate significant improvements over vanilla flow matching approaches, substantially narrowing the performance gap between generative segmentation and strong discriminative specialists. 
Project Page: \url{https://wang-chaoyang.github.io/project/flowseg}.
\end{abstract}
    
\begin{figure}[t]
    \centering
    \includegraphics[width=1.0\textwidth]{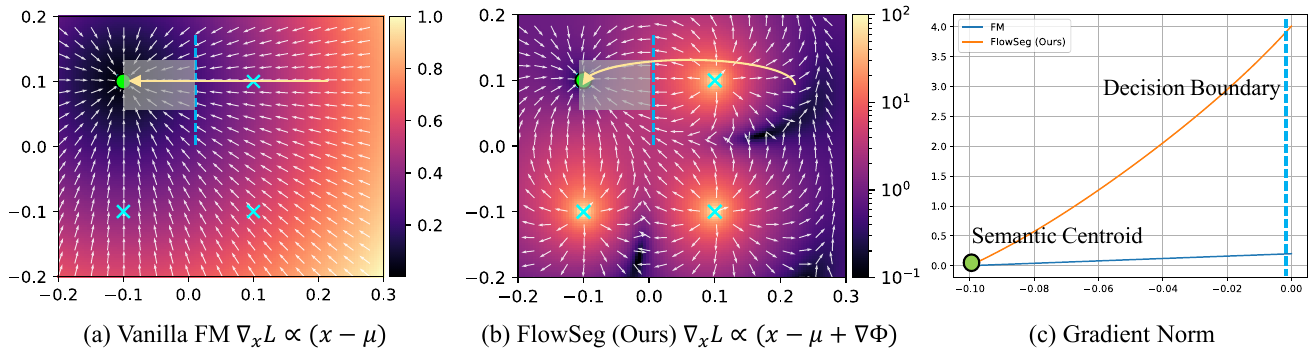}
    \caption{\textbf{Visualization of gradient vanishing and trajectory traversing in generative segmentation (x-prediction).}
(a) Vanilla flow matching suffers from vanishing gradients near semantic centroids $\mu$, resulting in slow and non-discriminative trajectories that may traverse proximal neighborhoods of competing centroids and cause false predictions.
(b) FlowSeg (ours) introduces a potential function $\Phi$ to enhance gradients around target centroids and enforce repulsion from non-targets, enabling faster convergence and more discriminative, deflected trajectories. The $x - \mu$ term maintains convergence near outer boundaries. 
(c) Gradient norm from centroid to decision boundary (gray regions in (a),(b)): Our method maintains strong gradients, whereas vanilla flow matching gradients nearly vanish. Yellow curves and blue dashed lines denote the predicted trajectory and decision boundary. The green dot marks the target centroid, and blue crosses indicate irrelevant categories.
}
    \label{fig:teaser}
\end{figure}

\section{Introduction}
\label{sec:intro}

Diffusion models~\cite{song2020denoising,song2020score,ho2020denoising} have emerged as a leading approach in visual generation, excelling at modeling complex data distributions. Their effectiveness in high-quality image synthesis~\cite{rombach2022high, podell2023sdxl,esser2024scaling}, video generation~\cite{blattmann2023stable,agarwal2025cosmos,hacohen2024ltx} and image editing~\cite{brooks2023instructpix2pix} has motivated extensions to various downstream tasks. While initially proposed for visual generation, recent works~\cite{yang2025vrmdiff,wang2024semflow,wang2024explore} have begun applying diffusion frameworks to perception tasks, aiming to unify generation and understanding within a common paradigm.
Among these tasks, segmentation is distinctive in that it assigns each pixel a discrete semantic label~\cite{long2015fully,chen2017deeplab,ronneberger2015u,zhao2017pyramid,cheng2021per,cheng2022masked}, producing categorical rather than continuous outputs. This poses a key challenge: diffusion models operate in continuous spaces, while segmentation requires inherently discrete predictions.
This gap introduces significant challenges in task formulation and optimization.

\begin{figure}[t!]
    \centering
    \includegraphics[width=0.8\textwidth]{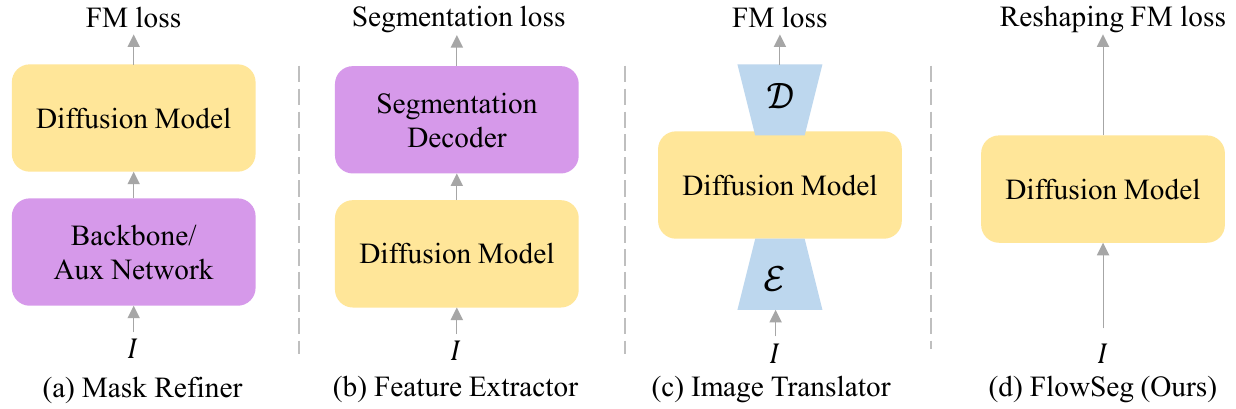}
    \caption{\textbf{Comparison of diffusion segmentation paradigms.}
(a) Diffusion models are used primarily for mask refinement, relying on an external backbone for feature extraction or auxiliary networks for coarse segmentation.
(b) Diffusion models serve as the backbone, followed by a dedicated segmentation head.
(c) The segmentation task is formulated as image-to-mask translation without auxiliary networks, yet still depends on a pretrained VAE.
(d) FlowSeg (ours) performs pixel-level end-to-end training without additional auxiliary modules, and rectifies vanilla flow matching by reshaping the underlying vector field for better optimization.
Noise is omitted for simplicity.
}
    \label{fig:method_cmp}
\end{figure}

To adapt diffusion models for segmentation and utilize their pretrained priors, existing methods fall into three main categories, as shown in Fig.~\ref{fig:method_cmp}.
1) Employing visual foundation features and adding lightweight diffusion modules to refine masks~\cite{wang2023segrefiner,geng2023instructdiffusion, fan2024toward}.
2) Using diffusion models as feature extractors with segmentation heads~\cite{zhao2023unleashing,xu2024matters,kondapaneni2024text,ji2023ddp,van2024simple,baranchuk2021label,khani2023slime}.
3) Treating segmentation as end-to-end image translation, encoding masks as three-channel pseudo-color images~\cite{amit2021segdiff,lee2024exploiting,wang2024explore,wang2024semflow,qi2024unigs,lin2024pixwizard}. 
Beyond model design, interestingly, semantic segmentation remains a challenging and underexplored task for diffusion-based models, primarily due to two factors: 1) the mismatch between the inherent stochasticity of generative models and the deterministic demands of semantic segmentation, and 2) the typically large number of classes involved.
Recent works~\cite{wang2024semflow, lee2024exploiting} address the first challenge by introducing deterministic flows, yet the issue of high class cardinality persists. To maintain tractability, existing approaches often simplify the problem setting. For instance, some methods reduce the class space by merging similar categories~\cite{xu2025jodi}, or reformulate the task as a series of referring image segmentation problems~\cite{geng2023instructdiffusion} to break it into simpler subproblems.
These findings indicate that stochasticity and high class cardinality are fundamental obstacles for diffusion segmentation, suggesting that the core difficulty lies not in model capacity, but in the optimization dynamics inherent to the framework.

In this paper, we analyze the issue from an optimization dynamics perspective and identify two key problems: \textit{gradient vanishing} and \textit{trajectory traversing}, as shown in Fig.~\ref{fig:teaser}. By examining the gradients induced by standard flow matching~\cite{lipman2022flow, liu2022rectified}, we observe that the optimization signal is directly scaled by the distance between predictions and semantic centroids. As class centroids reside in a bounded low-dimensional space, the gradients are inherently limited and decay rapidly as predictions approach the target, ultimately vanishing near convergence and causing optimization to stall.
Furthermore, the regression-based approach only provides attractive forces toward the ground-truth class, lacking explicit repulsion from competing centroids. This combination of weak gradients and absent inter-class repulsion results in slow convergence and poor semantic separation, ultimately degrading segmentation performance.

To overcome these limitations, we reshape the vector field by introducing a distance-aware correction term to the original velocity, enabling both attractive and repulsive interactions and sustaining non-vanishing gradients near centroids. This enhances class separation and accelerates convergence.
Additionally, we propose a quasi-random centroid encoding strategy inspired by Kronecker sequences, which is computationally efficient. Combined with an end-to-end pixel neural field framework, our design aligns the diffusion model with pixel-level tasks.

Our main contributions are summarized as follows: 
1) We analyze the slow convergence and suboptimal performance in diffusion segmentation, attributing them to gradient vanishing and trajectory traversing caused by the vanilla flow matching objective. 
2) We propose a novel vector field reshaping strategy that augments target velocity with a corrective term, preserving gradient magnitude and introducing explicit repulsion to improve semantic separation.
3) We employ a pixel neural field framework for end-to-end training and introduce a scalable, quasi-random centroid encoding scheme that ensures balanced inter-class geometry without extra optimization.
4) Extensive experiments show consistent and significant improvements, substantially narrowing the gap between generative and strong discriminative approaches.

\section{Related Work}
\label{sec:related_work}

\subsection{Diffusion and Flow Model}

Diffusion models~\cite{ho2020denoising} and flow models~\cite{lipman2022flow, liu2022flow} parameterize data generation as a time-dependent dynamical system that transports a simple reference distribution to a target data distribution.
They enable high-fidelity synthesis across diverse domains, including image generation~\cite{rombach2022high, podell2023sdxl,esser2024scaling}, video generation~\cite{blattmann2023stable,agarwal2025cosmos,hacohen2024ltx}, image editing~\cite{brooks2023instructpix2pix}, etc.
Early diffusion models typically use stochastic differential equations (SDEs) to learn score or noise estimates under a predefined Gaussian corruption process.
In contrast, flow models~\cite{liu2022flow, liu2023instaflow} directly learn a deterministic velocity field that defines a straight probability path, enabling more efficient sampling and greater flexibility in distribution transport.

To reduce computational overhead, conventional approaches~\cite{rombach2022high} predominantly operate within latent spaces compressed by Variational Autoencoders (VAEs)~\cite{kingma2013auto}. However, the inherent information loss in VAE compression often hinders tasks that require pixel-level precision and direct spatial alignment. To address this, recent pixel diffusion models~\cite{yu2025pixeldit, li2025back, wang2025pixnerd} have emerged, leveraging large patch decoding to maintain scalability without sacrificing fine-grained details. 
Despite differences in parameterization, these approaches share a common principle: generative modeling is reduced to learning a \textit{continuous} vector field that aligns probability trajectories between distributions, typically via regression-based objectives, which poses a challenge for extending to \textit{discrete} perception tasks.

\subsection{Generative Segmentation}

Visual segmentation aims to partition an image into semantically coherent regions with discrete class labels. While conventional methods~\cite{long2015fully,chen2017deeplab,ronneberger2015u,zhao2017pyramid,cheng2021per,cheng2022masked} typically frame segmentation as a discriminative dense classification task, generative formulations have recently gained traction for their superior uncertainty modeling and structured prediction capabilities~\cite{amit2021segdiff, liang2022gmmseg}. 

Existing diffusion segmentation methods generally follow three paradigms:
1) Refinement-based, which employ diffusion modules to iteratively polish masks generated by foundation models~\cite{wang2023segrefiner,geng2023instructdiffusion, fan2024toward};
2) Feature-based, which utilize diffusion backbones as powerful frozen extractors for task-specific segmentation heads~\cite{zhao2023unleashing,xu2024matters,kondapaneni2024text,ji2023ddp,van2024simple,baranchuk2021label,khani2023slime};
3) Unified-generative, which cast mask prediction as an end-to-end image translation process~\cite{amit2021segdiff,lee2024exploiting,wang2024explore,wang2024semflow,qi2024unigs,lin2024pixwizard}.

In the unified-generative formulation, segmentation masks are typically embedded into a low-dimensional continuous manifold to align with the continuous nature of generative dynamics. 
While such embedding is essential for generative modeling, simply adopting vanilla regression-based flow objectives fails to account for the discrete nature of semantic labels. This leads to an optimization mismatch where the vector field is learned as a generic density estimator, overlooking the categorical constraints inherent in segmentation.

\section{Optimization Dynamics Analysis}
\label{sec:analysis}

In this section, we provide a formal analysis of why the standard Mean Squared Error (MSE) objective, commonly used in Flow Matching for generative tasks, leads to suboptimal optimization dynamics and performance bottlenecks when applied to generative segmentation.

\subsection{Preliminary}
\label{sub_sec:preliminary}

Let $I \in \mathbb{R}^{H \times W \times 3}$ denote an input image. We define the segmentation task as a flow between a source distribution $p_0(x_0|I)$ and a target distribution $p_1(x_1)$. To eliminate the randomness of Gaussian noise and leverage the deterministic nature of segmentation, we define $x_0$ as a feature-dependent distribution (e.g., the input image itself or encoded image features) and $x_1$ as the target mask representation.

In a categorical or multi-channel segmentation setting, each pixel or region is mapped to a specific target centroid within a bounded continuous space (e.g., $\mathcal{C} \subset [-1, 1]^d$). Let $\mathcal{P} = \{\mu_1, \mu_2, \dots, \mu_N\}$ be the set of target centroids representing $N$ distinct categories or instances. In Rectified Flow~\cite{liu2022flow}, the probability path $x_t$ is constructed as follows:
\begin{equation}
x_t = (1-t)x_0 + t x_1, \quad t \in [0, 1]
\end{equation}
The corresponding ground-truth velocity (vector field) is:
\begin{equation}
v_{gt}(x_t) = \frac{dx_t}{dt} = x_1 - x_0
\end{equation}
The standard training objective is to learn a velocity field $v_\theta(x_t, t, I)$ parameterized by a neural network, which approximates the ground-truth via the MSE loss:
\begin{equation}
\mathcal{L}_{FM} = \mathbb{E}_{t, x_0, x_1} \left\lVert v_\theta(x_t, t, I) - v_{gt} \right\rVert^2
\end{equation}

\subsection{Gradient Analysis}

To understand the optimization challenges, we analyze the gradient of $\mathcal{L}_{FM}$ with respect to the network's prediction. The velocity $v_\theta$ implicitly defines a predicted target $\hat{x}_1$, which can be estimated as:
\begin{equation}
\hat{x}_1 = x_t + (1-t) v_\theta
\end{equation}
Substituting $v_\theta$ and $v_{gt}$ into the loss function, we can rewrite the objective in terms of the distance between the predicted state and the target centroid:

\begin{equation}
\mathcal{L}_{FM} \propto \left\lVert \frac{\hat{x}_1 - x_t}{1-t} - \frac{x_1 - x_t}{1-t} \right\rVert^2 = \frac{1}{(1-t)^2} \left\lVert \hat{x}_1 - x_1 \right\rVert^2
\end{equation}
By examining the gradient of this loss with respect to the predicted target $\hat{x}_1$, several critical issues emerge regarding the optimization dynamics:

\noindent
\textbf{Gradient Vanishing Near Centroids.}
The gradient magnitude is directly proportional to the Euclidean distance between the prediction and the target for fixed $t$:
\begin{equation}
\nabla_{\hat{x}_1} \mathcal{L}_{FM} \propto (\hat{x}_1 - x_1)
\end{equation}
As the prediction $\hat{x}_1$ approaches the target centroid $x_1 \in \mathcal{P}$, the gradient vanishes. In the late stages of training, the model lacks sufficient driving force to precisely converge to the exact centroid coordinates, leading to blurry segmentation boundaries. 

\noindent
\textbf{Limited Signal Scale in Bounded Space.}
Given that the centroids $\mathcal{P}$ are typically embedded in a compact space such as $[-1, 1]^d$, the maximum possible distance $\|\hat{x}_1 - x_1\|$ is intrinsically small. For tasks with high category cardinality $N$, the centroids are densely packed, and the MSE-based gradient provides a very weak signal for the model to distinguish between adjacent categorical anchors.

\noindent
\textbf{Absence of Repulsive Forces.}
The MSE loss is strictly a \textit{unimodal attraction} objective. It solely penalizes the distance to the target centroid and remains agnostic to competing centroids.
On the one hand, from the perspective of classical contrastive learning, a robust classifier should not only pull the prediction toward the positive category but also push it away from negative ones. 
On the other hand, from the perspective of flow matching, without a repulsive mechanism, target-bound trajectories may inadvertently \textit{traverse} the proximal neighborhoods of competing classes, inducing semantic ambiguity during the generative process.

\section{Method}
\label{sec:method}

In this section, we detail the proposed framework to rectify the optimization dynamics of generative segmentation. First, we introduce a deterministic, quasi-random scheme for category encoding. Then, we present the vector field reshaping mechanism, which is designed to introduce discriminative forces into the generative training objective. Finally, we propose using pixel neural field decoding for end-to-end generative segmentation training.

\subsection{Category Encoding}

To map $N$ semantic categories into the bounded 3D color space $\mathcal{C} \subset [-1, 1]^3$, we employ a Kronecker-style sequence, which is a quasi-random sequence characterized by determinism and low computational cost.

Let $\mathcal{V} = \{\sqrt{2}, \sqrt{3}, \sqrt{5}\}$ be a set of square roots of the first three primes. These values are linearly independent over the field of rational numbers $\mathbb{Q}$, ensuring that the generated sequence does not collapse onto lower-dimensional manifolds or exhibit periodic correlations. For each category index $k \in \{0, \dots, N-1\}$, the centroid $\mu_k$ is initially computed as:
\begin{equation}
\hat{c}_{k} = (k \cdot \mathcal{V}) \pmod{1}
\end{equation}
The resulting coordinates are then normalized and stretched to occupy the full volume of the $[-1, 1]^3$ cube:
\begin{equation}
\mu_k = 2 \cdot \text{Norm}(\hat{c}_k) - 1
\end{equation}
This approach ensures a relatively uniform distribution of centroids with high minimum inter-point distances, providing a stable and deterministic geometric basis for the vector field learning.

\subsection{Vector Field Reshaping}

As analyzed in Sec.~\ref{sec:analysis}, the vanilla flow matching objective fails to provide repulsive forces, leading to path traversing and gradient vanishing. To address this, we reshape the flow matching objective by augmenting the ground-truth velocity $v_{gt}$ with a discriminative rectification term.

\noindent
\textbf{Potential-based Vector Rectification.}
To instill discriminative awareness into the flow field, we construct a potential field $\Phi$ over the centroid space. Let $\hat{x}_1$ denote the estimated prediction, and $d_k = \|\hat{x}_1 - \mu_k\|^2$ be the squared Euclidean distance to target centroid $\mu_k \in \mathcal{P}$. We introduce a transformation operator $\mathcal{T}(\cdot)$ to map these geometric distances into a semantic embedding space. In our primary formulation, we define this operator as $\mathcal{T}(d) = -\log(d + \epsilon)$, where $\epsilon$ is to avoid zero-division. $\mathcal{T}$ remains flexible for various distance-warping strategies.

Distances are converted into a soft-assignment distribution $p$ via a temperature-scaled softmax over the transformed distances:
\begin{equation}
p_k = \frac{\exp(\mathcal{T}(d_k) / \tau)}{\sum_j \exp(\mathcal{T}(d_j) / \tau)}
\end{equation}
where $\tau$ is a temperature hyperparameter controlling the sharpness of the categorical assignment. To guide the optimization, we define the discriminative potential $\Phi$ as the divergence between the current assignment $p$ and the ground-truth one-hot distribution $y$:
\begin{equation}
\Phi(\hat{x}_1, y) = - \sum_{k=1}^N y_k \log p_k
\end{equation}
This potential reaches its global minimum when the prediction $\hat{x}_1$ perfectly aligns with the target centroid $\mu_{gt}$ while maximizing its separation from competitors. The rectification term $\nabla \Phi$ is obtained by computing the gradient of the potential with respect to the prediction $\hat{x}_1$. Utilizing the property $\mathcal{T}'(d) = - (d + \epsilon)^{-1}$, the gradient is formulated as:
\begin{equation}
\nabla \Phi = \frac{\partial \Phi}{\partial \hat{x}_1} = \frac{2}{\tau} \sum_{k=1}^N (y_k - p_k) \frac{\hat{x}_1 - \mu_k}{d_k + \epsilon}
\end{equation}

\noindent
\textbf{Objective Reformulation.}
The reshaped target velocity $\tilde{v}_t$ is defined by integrating the rectification term into the original velocity:
\begin{equation}
\label{eq:tilde_v}
\tilde{v}_t = v_t - \nabla \Phi
\end{equation}
To maintain the stability of the probability path and prevent the network from directly optimizing the potential field,
we apply the stop-gradient operator to the reshaped target. The final training objective is defined as:
\begin{equation}
\mathcal{L}_{res} =  \mathbb{E}_{t, x_0, x_1}\left\lVert v_\theta(x_t, t, I) - \text{sg}[\tilde{v}_t] \right\rVert^2
\end{equation}
where $\text{sg}[\cdot]$ denotes the detach operation.

\noindent
\textbf{Gradient Dynamics Comparison.}
To further elucidate the impact of Flow Reshaping, we compare the gradients of the standard Flow Matching loss and our proposed objective. For the sake of clarity, let $g_{\theta} = \nabla_{\theta} v_{\theta}$ denote the Jacobian of the network with respect to its parameters.

In the standard Flow Matching framework, the parameter update is driven by:
\begin{equation}
\nabla_{\theta} \mathcal{L}_{FM} = g_{\theta}^{\top} \left( v_{\theta} - (x_1 - x_0) \right)
\end{equation}
As $v_{\theta}$ approaches the constant target $v_{gt} = x_1 - x_0$, the error signal diminishes. This signal is purely reconstructive, providing no information regarding the proximity of the predicted flow to incorrect categorical centroids.

In contrast, our reshaped objective $\mathcal{L}_{res}$ yields the following gradient:
\begin{equation}
\nabla_{\theta} \mathcal{L}_{res} = g_{\theta}^{\top} \left( v_{\theta} - (x_1 - x_0 - \text{sg}[\nabla \Phi]) \right)
\end{equation}
By decomposing the term $\nabla \Phi$, we observe that a discriminative bias shifts the effective target for the network:
\begin{equation}
\text{Effective Target}
=
\underbrace{(x_1 - x_0)}_{\text{Reconstructive}}
+
\underbrace{
\frac{2}{\tau}
\sum_{k=1}^{N}
(p_k - y_k)
\frac{\hat{x}_1 - \mu_k}{
\left\lVert \hat{x}_1 - \mu_k \right\rVert^2 + \epsilon
}
}_{\text{Discriminative Correction}}
\end{equation}

\subsection{End-to-End Pixel Decoding}

Conventional VAE-based latent spaces are often ill-suited for high-precision segmentation due to \textit{manifold distortion} and \textit{optimization decoupling}. The former introduces latent artifacts that disrupt pixel-level details, while the latter prevents the diffusion process from aligning directly with segmentation objectives. Moreover, the high memory overhead of VAEs makes full end-to-end optimization computationally infeasible.

To circumvent these limitations, following PixNerd~\cite{wang2025pixnerd}, we propose using pixel neural field decoding for end-to-end generative segmentation training. Instead of decoding a patch feature $\mathbf{X}^n \in \mathbb{R}^D$ through a simplistic linear projection, we treat each patch as a continuous neural field. The Transformer backbone serves as a parameter generator, predicting the weights of a localized Multi-Layer Perceptron (MLP).
For the $n$-th patch, the network dynamically predicts the weight matrices $\{\mathbf{W}_1^n, \mathbf{W}_2^n\}$ of a lightweight MLP from the patch feature. These weights are generated via a projection and subsequently normalized to ensure numerical stability, yielding the normalized weights $\bar{\mathbf{W}}_1^n$ and $\bar{\mathbf{W}}_2^n$:
$$
\mathbf{W}_1^n, \mathbf{W}_2^n = \text{Linear}(\mathbf{X}^n)
$$
To query the velocity field at any spatial coordinate $(i, j)$ within the patch, we construct an input context $\mathbf{h}_{i,j}$ by concatenating the Discrete Cosine Transform (DCT) positional encoding with the noisy pixel state $\mathbf{x}_t(i, j)$:
\begin{equation*}
\mathbf{h}_{i,j} = \text{Concat}([\text{DCT}(i, j), x_t(i, j)])
\end{equation*}
The final pixel-wise velocity $v^n(i, j)$ is then directly decoded by applying the dynamically generated MLP:
$$
v^n(i, j) = \bar{\mathbf{W}}_2^n \sigma(\bar{\mathbf{W}}_1^n \mathbf{h}_{i,j})
$$
where $\sigma$ denotes the SiLU activation function.

\section{Experiment}
\label{sec:exp}

In this section, we first detail our experimental settings and benchmark our approach against baseline methods both quantitatively and qualitatively. Finally, we present comprehensive ablation studies dissecting our macroscopic modeling choices, objective formulations, operator dynamics, and state representations.

\subsection{Experimental Settings}

\noindent 
\textbf{Dataset.} 
We evaluate our approach on two high-cardinality datasets: ADE20K~\cite{zhou2017scene} and COCO-Stuff~\cite{caesar2018coco}. ADE20K is a scene parsing dataset covering 150 fine-grained semantic concepts consisting
of 20k images. COCO-Stuff covers 171 labels and consists of
164k images.

\noindent
\textbf{Implementation.}
Our model is initialized with PixNerd~\cite{wang2025pixnerd} weights. All images and segmentation masks are resized to a uniform resolution. Training proceeds in two progressive stages, using the AdamW optimizer and Representation Alignment (REPA)~\cite{yu2024representation} to accelerate convergence. Stage 1 uses a $256\times 256$ resolution with a batch size of 64, followed by Stage 2 at $512\times 512$ with a batch size of 16. The default temperature is 1.0. In line with standard practice, binary masks are applied to filter out unlabeled regions, which are omitted from our notation for brevity.

\noindent
\textbf{Baselines.}
We consider discriminative specialists and diffusion models. For the former, we include DeeplabV3+~\cite{chen2018encoder}, OCRNet~\cite{yuan2020object}, SegFormer~\cite{xie2021segformer} and MaskFormer~\cite{cheng2021per}. 
For the latter, we include InstructDiffusion~\cite{geng2023instructdiffusion}, PixWizard~\cite{lin2024pixwizard} and SymmFlow~\cite{caetano2025symmetrical}. 

\subsection{Main Results}

\begin{table}[t]
\centering
\caption{\textbf{Comparisons on ADE20K and COCO-Stuff dataset.} $\dagger$ indicates that it performs referring image segmentation, which requires predicting a separate mask for each semantic category. LSTI and  ($\cdot$) indicate large-scale text-image datasets and the initialized weight, respectively. \textbf{Bold} and \underline{underline} indicate the first and second best entries.}
\begin{subtable}{0.48\textwidth}
\centering
\caption{Performance on ADE20K dataset}
\label{tab:main_ade}
\resizebox{\linewidth}{!}{%
\begin{tabular}{lccc}
\toprule[0.1em]
Method & Backbone & Pretrain Data & mIoU \\
\midrule[0.1em]
\multicolumn{4}{l}{\textcolor{gray}{\textit{Discriminative Specialists}}} \\
DeeplabV3+ & ResNet101 & IN-1k & 44.1 \\
SegFormer & MiT-B2 & IN-1k & 46.5 \\
MaskFormer & Swin-T & IN-1k & \underline{46.7} \\
\hline
\multicolumn{4}{l}{\textcolor{gray}{\textit{Diffusion Models}}} \\
InstructDiffusion$^\dagger$ & (SD1.5) & LSTI & 33.6 \\
PixWizard & (Lumina-Next-T2I) & LSTI & 32.8 \\
FlowSeg (Ours) & (PixNerd) & IN-1k & \textbf{47.1} \\
\bottomrule[0.1em]
\end{tabular}%
}
\end{subtable}%
\hfill
\begin{subtable}{0.48\textwidth}
\centering
\caption{Performance on COCO-Stuff dataset}
\label{tab:main_cocostuff}
\resizebox{\linewidth}{!}{%
\begin{tabular}{lccc}
\toprule[0.1em]
Method & Backbone & Pretrain Data & mIoU \\
\midrule[0.1em]
\multicolumn{4}{l}{\textcolor{gray}{\textit{Discriminative Specialists}}} \\
DeeplabV3+ & ResNet50 & IN-1k & 38.4 \\
OCRNet & HRNet-W48 & IN-1k & 42.3 \\
SegFormer & MiT-B2 & IN-1k & \underline{44.6} \\
\hline
\multicolumn{4}{l}{\textcolor{gray}{\textit{Diffusion Models}}} \\
SymmFlow & (SD2.1) & LSTI & 39.6 \\
FlowSeg (Ours) & (PixNerd) & IN-1k & \textbf{44.9} \\
\bottomrule[0.1em]
\end{tabular}%
}
\end{subtable}
\end{table}

\noindent
\textbf{Quantitative Results.}
Tab.~\ref{tab:main_ade} and Tab.~\ref{tab:main_cocostuff} benchmark FlowSeg against discriminative specialists and diffusion-based models. Existing diffusion paradigms lag significantly behind discriminative baselines, even when leveraging the extensive priors of large-scale pretrained models like Stable Diffusion~\cite{rombach2022high}. This performance gap corroborates our analysis in Sec.~\ref{sec:analysis} regarding the optimization bottlenecks inherent to generative segmentation. Specifically, on ADE20K, InstructDiffusion and PixWizard trail DeepLabV3+ by over 10 mIoU. Similarly, on COCO-Stuff, SymmFlow underperforms SegFormer by 5 mIoU despite utilizing SD2.1 weights. In contrast, empowered by the reshaped vector field, FlowSeg achieves 47.1 mIoU on ADE20K and 44.9 mIoU on COCO-Stuff. Notably, FlowSeg successfully surpasses strong discriminative specialists solely relying on IN-1k pretraining.

\begin{figure}[t]
	\centering
	\includegraphics[width=0.75\linewidth]{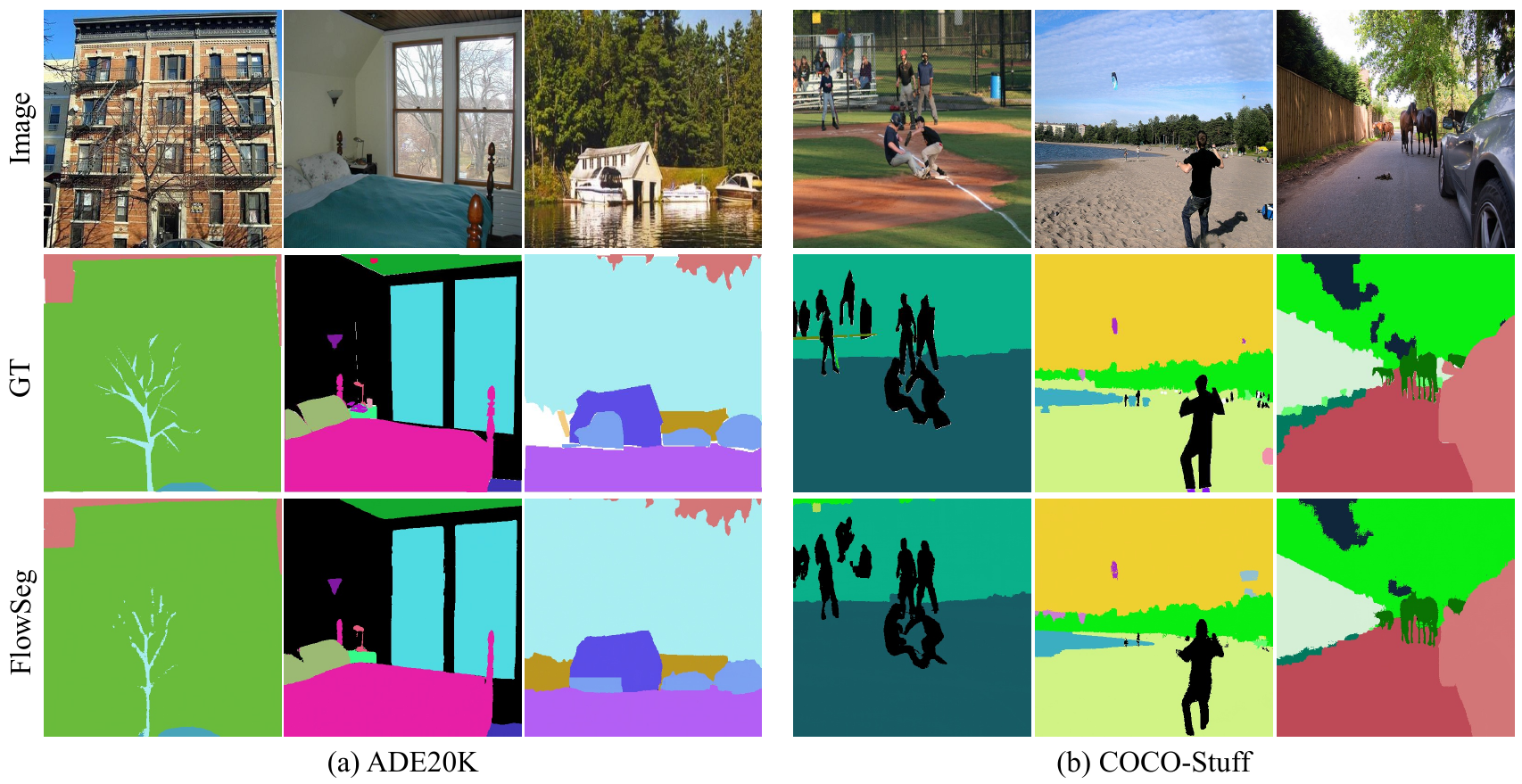}
	\caption{Visualization of segmentation results on (a) ADE20K and (b) COCO-Stuff datasets. Color white in the ground truth (GT) denotes the ignored regions. As ADE20K and COCO-Stuff datasets have different category cardinality, the same color between (a) and (b) does not necessarily represent the same semantic category.}
    \label{fig:main_vis}
\end{figure}

\begin{figure}[t]
	\centering
	\includegraphics[width=1.0\linewidth]{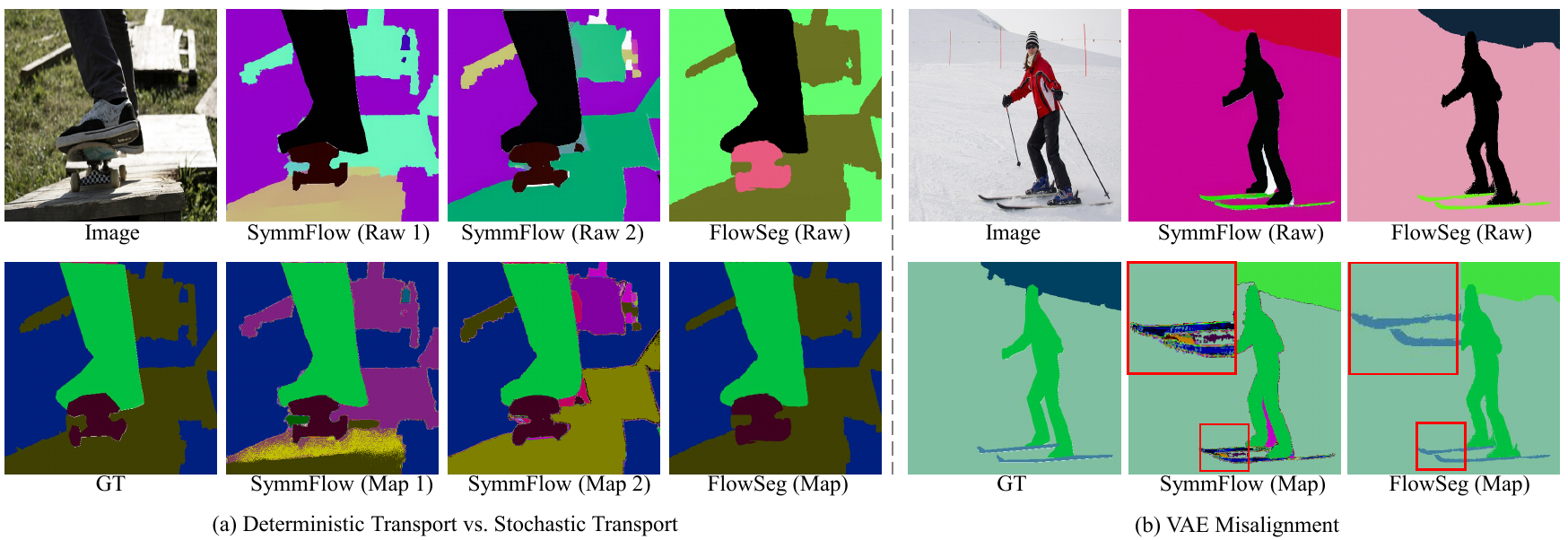}
	\caption{\textbf{Visual comparisons between FlowSeg (ours) and SymmFlow (Baseline).} The diffusion model first predicts pseudo-masks (\textbf{Raw}), then maps them to the nearest semantic centroids to obtain the final masks (\textbf{Map}). (a) Comparison between deterministic (ours) and stochastic modeling: SymmFlow’s predictions vary with random seeds, while ours remain consistent. (b) VAE-based latent space modeling produces masks with similar colors that may not correspond to the correct semantic categories, due to imperfect alignment with pixel-level centroids.}
    \label{fig:abla_symm}
\end{figure}

\noindent
\textbf{Qualitative Results.}
Fig.~\ref{fig:main_vis} demonstrates FlowSeg's robust segmentation capabilities across diverse and complex scenarios, effectively resolving varying object scales in both indoor and outdoor environments. Furthermore, we contrast FlowSeg with SymmFlow, which maps the joint distribution of images and Gaussian noise to segmentation masks. As shown in Fig.~\ref{fig:abla_symm}a, SymmFlow's stochastic formulation fails to yield deterministic predictions, leading to inconsistent outputs. Conversely, FlowSeg strictly enforces deterministic transport, yielding qualitatively superior and highly consistent semantic boundaries.

\begin{figure}[t]
	\centering
	\includegraphics[width=1.0\linewidth]{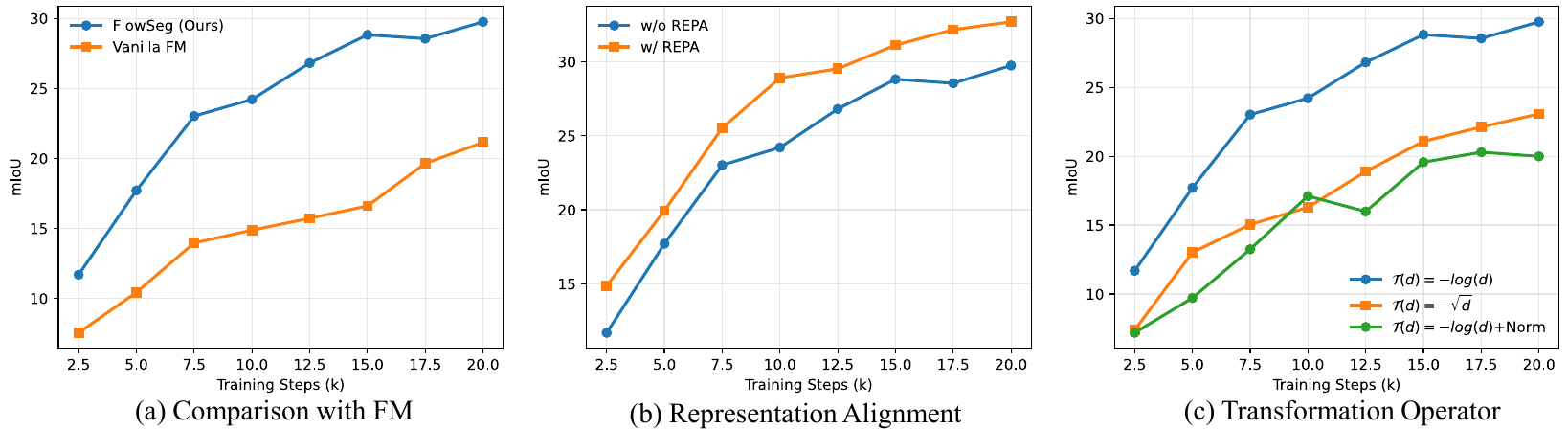}
	\caption{Convergence comparison of different training recipes. (a) FlowSeg vs. vanilla flow matching. (b) Training w/ REPA vs. w/o REPA, (c) Different transformation operators $\mathcal{T}$.}
    \label{fig:abla_converge}
\end{figure}

\begin{table*}[ht]
\centering
\small
\setlength{\tabcolsep}{3pt}
\caption{\textbf{Ablations on training, representations, and operators.} (a) Mask excludes unlabeled regions (defaults use v-pred). (b) T-comp scales x-pred loss by $(1-t)^2$ to offset time dependence. (c) Concat conditions images with padded maps (Gaussian/Zero initialized) under vanilla flow matching. $K$ denotes category cardinality.}
\label{tab:abla_design}
\label{tab:abla_dim_op}
\begin{subtable}{0.20\textwidth}
\centering
\caption{Design choices}
\label{tab:abla_design_sub}
\begin{tabular}{ccc}
\toprule[0.1em]
\makecell{Pred.} & Mask & mIoU \\
\midrule[0.1em]
v-pred &             & 28.15 \\
v-pred & \checkmark  & 29.75 \\
x-pred & \checkmark  & 29.06 \\
\bottomrule[0.1em]
\end{tabular}
\end{subtable}%
\hfill
\begin{subtable}{0.22\textwidth}
\centering
\caption{Time compensation}
\label{tab:abla_tcomp}
\begin{tabular}{ccc}
\toprule[0.1em]
\makecell{Reshape\\Field} & \makecell{T-\\comp.} & mIoU \\
\midrule[0.1em]
             &              & 22.05 \\
             & \checkmark   & 18.30 \\
\checkmark   &              & 29.06 \\
\checkmark   & \checkmark   & 30.68 \\
\bottomrule[0.1em]
\end{tabular}
\end{subtable}%
\hfill
\begin{subtable}{0.26\textwidth}
\centering
\caption{Dimension ablation}
\label{tab:abla_dim}
\begin{tabular}{cccc}
\toprule[0.1em]
Dim & Concat & Init & mIoU \\
\midrule[0.1em]
3   & -          & -     & 20.60 \\
3   & \checkmark & Gauss & 5.03  \\
3   & \checkmark & Zero  & 17.28 \\
$K$ & \checkmark & Gauss & 0.29  \\
$K$ & \checkmark & Zero  & 1.02  \\
\bottomrule[0.1em]
\end{tabular}
\end{subtable}%
\hfill
\begin{subtable}{0.26\textwidth}
\centering
\caption{Operator ablation}
\label{tab:abla_op}
\begin{tabular}{cc}
\toprule[0.1em]
\makecell{Oper.} & mIoU \\
\midrule[0.1em]
$-\log(d)$           & 29.75 \\
$1/(d+\epsilon)$     & 5.46  \\
$1/\sqrt{d}$         & 23.44 \\
$-\log(d)-\sqrt{d}$ & 30.73 \\ 
\bottomrule[0.1em]
\end{tabular}
\end{subtable}
\end{table*}

\subsection{Ablation Studies}

\noindent
\textbf{Macroscopic Modeling.}
We validate our macroscopic design in Fig.~\ref{fig:abla_symm} and Fig.~\ref{fig:abla_converge}.
Unlike SymmFlow, which relies on Gaussian noise and yields unstable predictions across random seeds in Fig.~\ref{fig:abla_symm}a, FlowSeg strictly leverages a deterministic flow to meet semantic segmentation requirements.
Furthermore, traditional VAE-based methods struggle with pixel-wise precision due to latent space distortion. In contrast, our pixel-level end-to-end modeling reliably distinguishes fine-grained semantics in Fig.~\ref{fig:abla_symm}b.
This architecture also guarantees superior convergence. Fig.~\ref{fig:abla_converge}a demonstrates FlowSeg converges significantly faster than vanilla flow matching from the outset. Incorporating REPA further accelerates this process, boosting mIoU from 29.8 to 32.7 in Fig.~\ref{fig:abla_converge}b.

\noindent
\textbf{Training Objectives.}
Tab.~\ref{tab:abla_design_sub} and Tab.~\ref{tab:abla_tcomp} detail our objective formulation.
First, excluding unlabeled pixels effectively prevents semantically ambiguous areas from disrupting optimization, improving the v-prediction baseline from 28.15 to 29.75 mIoU in Tab.~\ref{tab:abla_design_sub}. We also observe a modest performance gap between v-prediction and x-prediction.
Second, Tab.~\ref{tab:abla_tcomp} reveals that the failure of vanilla x-prediction is intrinsic to transport field geometry. Explicit time compensation via $(1-t)^2$ fails to rescue vanilla flow matching, dropping performance from 22.05 to 18.30 mIoU. Conversely, our reshape field substantially boosts x-prediction to 29.06 mIoU, peaking at 30.68 mIoU with time compensation.

\textbf{Operator Dynamics.}
The geometric reshaping strictly relies on the transformation operator $\mathcal{T}$, which dictates the proxy gradient by mapping squared Euclidean distances $d$ to class posteriors via softmax. Analyzed in Tab.~\ref{tab:abla_op} and Fig.~\ref{fig:abla_converge}c, an effective operator must drive the target logit to strongly dominate, suppressing incorrect categories by creating a sharp probability peak and a sufficient margin. As $d \to 0$, the logarithmic term $-\log(d) \to +\infty$, providing a natural singularity with a manageable gradient proportional to $1/d$. This avoids the explosive instability of the hyper-singular $1/(d+\epsilon)$, which collapses to 5.46 mIoU. Conversely, milder operators like $1/\sqrt{d}$ (23.44 mIoU) exhibit sluggish convergence due to weak far-field shaping. The optimal formulation, $-\log(d)-\sqrt{d}$ (30.73 mIoU), seamlessly bridges near-centroid singularity with a far-field distance penalty. Ultimately, a robust operator must simultaneously enforce near-centroid attraction, wrong-centroid suppression, and stable global gradients.

\noindent
\textbf{Representation and Initialization.}
Tab.~\ref{tab:abla_dim} evaluates the flow state space.
The compact 3-channel cube drastically outperforms the $K$-channel one-hot embedding, yielding 17.28 versus 1.02 mIoU under zero initialization. One-hot targets reside on sparse simplex vertices. When paired with a Gaussian source, the noise spans the entire $\mathbb{R}^K$ space. For high cardinality (e.g., $K=150$), this injects massive off-simplex perturbations along semantically meaningless directions, forcing the model into irrelevant high-dimensional denoising. Since segmentation demands deterministic dense prediction rather than inherently stochastic generation, these unconstrained perturbations severely disrupt the transport mapping.
Consequently, zero initialization strictly surpasses Gaussian initialization. While Gaussian noise also degrades the 3-channel variant (dropping from 17.28 to 5.03 mIoU), the lower-dimensional state space reduces the expected perturbation magnitude. In contrast, high-dimensional $K$-channel Gaussian noise thoroughly derails optimization by dragging trajectories into non-semantic spaces.

\section{Conclusion}
\label{sec:conclusion}

This paper explores generative segmentation via flow matching, particularly in high-cardinality segmentation scenarios. We identified two key obstacles to previous methods: gradient vanishing and trajectory traversing. However, these problems are further obscured and entangled by classical latent-space modeling. 
To tackle these issues, we propose an end-to-end generative segmentation pipeline based on a pixel neural field. Then, we propose a novel vector field reshaping approach that simultaneously solves both problems while maintaining a high gradient near semantic centroids and introducing repulsion in the velocity field. 
Empirical studies and comprehensive ablations demonstrate significant improvements in convergence speed and final performance over classic flow-matching training recipes, and they closely approach strong discriminative specialists. We hope our research can inspire the community to rethink generative segmentation.

\noindent
\textbf{Acknowledgment.} This work was supported by NSFC (Grant No.42271426) and Shenzhen Science and Technology Innovation Project (Grant No.2025110105).

{\small
\bibliographystyle{ieee_fullname}
\bibliography{main}

@String(CVPR  = {IEEE Conf. Comput. Vis. Pattern Recog.})

@String(ICCV  = {Int. Conf. Comput. Vis.})

@String(ECCV  = {Eur. Conf. Comput. Vis.})

@String(NeurIPS = {Adv. Neural Inform. Process. Syst.})

@String(ICLR  = {Int. Conf. Learn. Represent.})

@String(AAAI  = {AAAI})

@String(CVPR  = {CVPR})

@String(ICCV  = {ICCV})

@String(ECCV  = {ECCV})

@String(NeurIPS = {NeurIPS})

@String(ICLR  = {ICLR})

@inproceedings{song2020denoising,
  title={Denoising diffusion implicit models},
  author={Song, Jiaming and Meng, Chenlin and Ermon, Stefano},
  booktitle={ICLR},
  year={2021}
}

@inproceedings{song2020score,
  title={Score-based generative modeling through stochastic differential equations},
  author={Song, Yang and Sohl-Dickstein, Jascha and Kingma, Diederik P and Kumar, Abhishek and Ermon, Stefano and Poole, Ben},
  booktitle={ICLR},
  year={2021}
}

@inproceedings{rombach2022high,
  title={High-resolution image synthesis with latent diffusion models},
  author={Rombach, Robin and Blattmann, Andreas and Lorenz, Dominik and Esser, Patrick and Ommer, Bj{\"o}rn},
  booktitle={CVPR},
  year={2022}
}

@inproceedings{zhao2023unleashing,
  title={Unleashing text-to-image diffusion models for visual perception},
  author={Zhao, Wenliang and Rao, Yongming and Liu, Zuyan and Liu, Benlin and Zhou, Jie and Lu, Jiwen},
  booktitle={ICCV},
  year={2023}
}

@inproceedings{geng2023instructdiffusion,
  title={Instructdiffusion: A generalist modeling interface for vision tasks},
  author={Geng, Zigang and Yang, Binxin and Hang, Tiankai and Li, Chen and Gu, Shuyang and Zhang, Ting and Bao, Jianmin and Zhang, Zheng and Li, Houqiang and Hu, Han and Chen, Dong and Guo, Baining},
  booktitle={CVPR},
  year={2024}
}

@inproceedings{brooks2023instructpix2pix,
  title={Instructpix2pix: Learning to follow image editing instructions},
  author={Brooks, Tim and Holynski, Aleksander and Efros, Alexei A},
  booktitle={CVPR},
  year={2023}
}

@inproceedings{podell2023sdxl,
  title={Sdxl: Improving latent diffusion models for high-resolution image synthesis},
  author={Podell, Dustin and English, Zion and Lacey, Kyle and Blattmann, Andreas and Dockhorn, Tim and M{\"u}ller, Jonas and Penna, Joe and Rombach, Robin},
  booktitle={ICLR},
  year={2024}
}

@article{amit2021segdiff,
  title={Segdiff: Image segmentation with diffusion probabilistic models},
  author={Amit, Tomer and Shaharbany, Tal and Nachmani, Eliya and Wolf, Lior},
  journal={arXiv preprint arXiv:2112.00390},
  year={2021}
}

@article{van2024simple,
  title={A Simple Latent Diffusion Approach for Panoptic Segmentation and Mask Inpainting},
  author={Van Gansbeke, Wouter and De Brabandere, Bert},
  journal={arXiv preprint arXiv:2401.10227},
  year={2024}
}

@inproceedings{long2015fully,
  title={Fully convolutional networks for semantic segmentation},
  author={Long, Jonathan and Shelhamer, Evan and Darrell, Trevor},
  booktitle={CVPR},
  year={2015}
}

@article{chen2017deeplab,
  title={Deeplab: Semantic image segmentation with deep convolutional nets, atrous convolution, and fully connected crfs},
  author={Chen, Liang-Chieh and Papandreou, George and Kokkinos, Iasonas and Murphy, Kevin and Yuille, Alan L},
  journal={TPAMI},
  year={2017},
}

@inproceedings{ronneberger2015u,
  title={U-net: Convolutional networks for biomedical image segmentation},
  author={Ronneberger, Olaf and Fischer, Philipp and Brox, Thomas},
  booktitle={MICCAI},
  year={2015},
}

@inproceedings{zhao2017pyramid,
  title={Pyramid scene parsing network},
  author={Zhao, Hengshuang and Shi, Jianping and Qi, Xiaojuan and Wang, Xiaogang and Jia, Jiaya},
  booktitle={CVPR},
  year={2017}
}

@inproceedings{cheng2021per,
  title={Per-pixel classification is not all you need for semantic segmentation},
  author={Cheng, Bowen and Schwing, Alex and Kirillov, Alexander},
  booktitle={NeurIPS},
  year={2021}
}

@inproceedings{cheng2022masked,
  title={Masked-attention mask transformer for universal image segmentation},
  author={Cheng, Bowen and Misra, Ishan and Schwing, Alexander G and Kirillov, Alexander and Girdhar, Rohit},
  booktitle={CVPR},
  year={2022}
}

@inproceedings{ho2020denoising,
  title={Denoising diffusion probabilistic models},
  author={Ho, Jonathan and Jain, Ajay and Abbeel, Pieter},
  booktitle={NeurIPS},
  year={2020}
}

@inproceedings{qi2024unigs,
  title={UniGS: Unified Representation for Image Generation and Segmentation},
  author={Qi, Lu and Yang, Lehan and Guo, Weidong and Xu, Yu and Du, Bo and Jampani, Varun and Yang, Ming-Hsuan},
  booktitle={CVPR},
  year={2024}
}

@inproceedings{liu2022flow,
  title={Flow straight and fast: Learning to generate and transfer data with rectified flow},
  author={Liu, Xingchao and Gong, Chengyue and Liu, Qiang},
  booktitle={ICLR},
  year={2023}
}

@inproceedings{liu2023instaflow,
  title={Instaflow: One step is enough for high-quality diffusion-based text-to-image generation},
  author={Liu, Xingchao and Zhang, Xiwen and Ma, Jianzhu and Peng, Jian and others},
  booktitle={ICLR},
  year={2024}
}

@article{esser2024scaling,
  title={Scaling rectified flow transformers for high-resolution image synthesis},
  author={Esser, Patrick and Kulal, Sumith and Blattmann, Andreas and Entezari, Rahim and M{\"u}ller, Jonas and Saini, Harry and Levi, Yam and Lorenz, Dominik and Sauer, Axel and Boesel, Frederic and others},
  journal={arXiv preprint arXiv:2403.03206},
  year={2024}
}

@article{liu2022rectified,
  title={Rectified flow: A marginal preserving approach to optimal transport},
  author={Liu, Qiang},
  journal={arXiv preprint arXiv:2209.14577},
  year={2022}
}

@inproceedings{lipman2022flow,
  title={Flow matching for generative modeling},
  author={Lipman, Yaron and Chen, Ricky TQ and Ben-Hamu, Heli and Nickel, Maximilian and Le, Matt},
  booktitle={ICLR},
  year={2023}
}

@article{wang2024explore,
  title={Explore in-context segmentation via latent diffusion models},
  author={Wang, Chaoyang and Li, Xiangtai and Ding, Henghui and Qi, Lu and Zhang, Jiangning and Tong, Yunhai and Loy, Chen Change and Yan, Shuicheng},
  journal={arXiv preprint arXiv:2403.09616},
  year={2024}
}

@inproceedings{caetano2025symmetrical,
  title={Symmetrical Flow Matching: Unified Image Generation, Segmentation, and Classification with Score-Based Generative Models},
  author={Caetano, Francisco and Viviers, Christiaan and De With, Peter HN and van der Sommen, Fons},
  booktitle={AAAI},
  year={2026}
}

@inproceedings{xie2021segformer,
  title={SegFormer: Simple and efficient design for semantic segmentation with transformers},
  author={Xie, Enze and Wang, Wenhai and Yu, Zhiding and Anandkumar, Anima and Alvarez, Jose M and Luo, Ping},
  booktitle={NeurIPS},
  year={2021}
}

@inproceedings{chen2018encoder,
  title={Encoder-decoder with atrous separable convolution for semantic image segmentation},
  author={Chen, Liang-Chieh and Zhu, Yukun and Papandreou, George and Schroff, Florian and Adam, Hartwig},
  booktitle={ECCV},
  year={2018}
}

@inproceedings{yuan2020object,
  title={Object-contextual representations for semantic segmentation},
  author={Yuan, Yuhui and Chen, Xilin and Wang, Jingdong},
  booktitle={ECCV},
  year={2020},
}

@inproceedings{lin2024pixwizard,
  title={Pixwizard: Versatile image-to-image visual assistant with open-language instructions},
  author={Lin, Weifeng and Wei, Xinyu and Zhang, Renrui and Zhuo, Le and Zhao, Shitian and Huang, Siyuan and Xie, Junlin and Gao, Peng and Li, Hongsheng},
  booktitle={ICLR},
  year={2025}
}

@inproceedings{zhou2017scene,
  title={Scene parsing through ade20k dataset},
  author={Zhou, Bolei and Zhao, Hang and Puig, Xavier and Fidler, Sanja and Barriuso, Adela and Torralba, Antonio},
  booktitle={CVPR},
  year={2017}
}

@inproceedings{caesar2018coco,
  title={Coco-stuff: Thing and stuff classes in context},
  author={Caesar, Holger and Uijlings, Jasper and Ferrari, Vittorio},
  booktitle={CVPR},
  year={2018}
}

@inproceedings{yu2024representation,
  title={Representation alignment for generation: Training diffusion transformers is easier than you think},
  author={Yu, Sihyun and Kwak, Sangkyung and Jang, Huiwon and Jeong, Jongheon and Huang, Jonathan and Shin, Jinwoo and Xie, Saining},
  booktitle={ICLR},
  year={2025}
}

@article{yu2025pixeldit,
  title={Pixeldit: Pixel diffusion transformers for image generation},
  author={Yu, Yongsheng and Xiong, Wei and Nie, Weili and Sheng, Yichen and Liu, Shiqiu and Luo, Jiebo},
  journal={arXiv preprint arXiv:2511.20645},
  year={2025}
}

@article{li2025back,
  title={Back to basics: Let denoising generative models denoise},
  author={Li, Tianhong and He, Kaiming},
  journal={arXiv preprint arXiv:2511.13720},
  year={2025}
}

@article{wang2025pixnerd,
  title={Pixnerd: Pixel neural field diffusion},
  author={Wang, Shuai and Gao, Ziteng and Zhu, Chenhui and Huang, Weilin and Wang, Limin},
  journal={arXiv preprint arXiv:2507.23268},
  year={2025}
}

@article{kingma2013auto,
  title={Auto-encoding variational bayes},
  author={Kingma, Diederik P and Welling, Max},
  journal={arXiv preprint arXiv:1312.6114},
  year={2013}
}

@article{blattmann2023stable,
  title={Stable video diffusion: Scaling latent video diffusion models to large datasets},
  author={Blattmann, Andreas and Dockhorn, Tim and Kulal, Sumith and Mendelevitch, Daniel and Kilian, Maciej and Lorenz, Dominik and Levi, Yam and English, Zion and Voleti, Vikram and Letts, Adam and others},
  journal={arXiv preprint arXiv:2311.15127},
  year={2023}
}

@article{agarwal2025cosmos,
  title={Cosmos world foundation model platform for physical ai},
  author={Agarwal, Niket and Ali, Arslan and Bala, Maciej and Balaji, Yogesh and Barker, Erik and Cai, Tiffany and Chattopadhyay, Prithvijit and Chen, Yongxin and Cui, Yin and Ding, Yifan and others},
  journal={arXiv preprint arXiv:2501.03575},
  year={2025}
}

@article{hacohen2024ltx,
  title={Ltx-video: Realtime video latent diffusion},
  author={HaCohen, Yoav and Chiprut, Nisan and Brazowski, Benny and Shalem, Daniel and Moshe, Dudu and Richardson, Eitan and Levin, Eran and Shiran, Guy and Zabari, Nir and Gordon, Ori and others},
  journal={arXiv preprint arXiv:2501.00103},
  year={2024}
}

@inproceedings{wang2024semflow,
  title={Semflow: Binding semantic segmentation and image synthesis via rectified flow},
  author={Wang, Chaoyang and Li, Xiangtai and Qi, Lu and Ding, Henghui and Tong, Yunhai and Yang, Ming-Hsuan},
  booktitle={NeurIPS},
  year={2024}
}

@article{xu2024matters,
  title={What matters when repurposing diffusion models for general dense perception tasks?},
  author={Xu, Guangkai and Ge, Yongtao and Liu, Mingyu and Fan, Chengxiang and Xie, Kangyang and Zhao, Zhiyue and Chen, Hao and Shen, Chunhua},
  journal={arXiv preprint arXiv:2403.06090},
  year={2024}
}

@inproceedings{kondapaneni2024text,
  title={Text-image alignment for diffusion-based perception},
  author={Kondapaneni, Neehar and Marks, Markus and Knott, Manuel and Guimaraes, Rog{\'e}rio and Perona, Pietro},
  booktitle={CVPR},
  year={2024}
}

@inproceedings{ji2023ddp,
  title={Ddp: Diffusion model for dense visual prediction},
  author={Ji, Yuanfeng and Chen, Zhe and Xie, Enze and Hong, Lanqing and Liu, Xihui and Liu, Zhaoqiang and Lu, Tong and Li, Zhenguo and Luo, Ping},
  booktitle={ICCV},
  year={2023}
}

@inproceedings{lee2024exploiting,
  title={Exploiting diffusion prior for generalizable dense prediction},
  author={Lee, Hsin-Ying and Tseng, Hung-Yu and Lee, Hsin-Ying and Yang, Ming-Hsuan},
  booktitle={CVPR},
  year={2024}
}

@inproceedings{baranchuk2021label,
  title={Label-efficient semantic segmentation with diffusion models},
  author={Baranchuk, Dmitry and Rubachev, Ivan and Voynov, Andrey and Khrulkov, Valentin and Babenko, Artem},
  booktitle={ICLR},
  year={2022}
}

@inproceedings{khani2023slime,
  title={Slime: Segment like me},
  author={Khani, Aliasghar and Taghanaki, Saeid Asgari and Sanghi, Aditya and Amiri, Ali Mahdavi and Hamarneh, Ghassan},
  booktitle={ICLR},
  year={2024}
}

@inproceedings{wang2023segrefiner,
  title={Segrefiner: Towards model-agnostic segmentation refinement with discrete diffusion process},
  author={Wang, Mengyu and Ding, Henghui and Liew, Jun Hao and Liu, Jiajun and Zhao, Yao and Wei, Yunchao},
  booktitle={NeurIPS},
  year={2023}
}

@article{fan2024toward,
  title={Toward a diffusion-based generalist for dense vision tasks},
  author={Fan, Yue and Xian, Yongqin and Zhai, Xiaohua and Kolesnikov, Alexander and Naeem, Muhammad Ferjad and Schiele, Bernt and Tombari, Federico},
  journal={arXiv preprint arXiv:2407.00503},
  year={2024}
}

@inproceedings{liang2022gmmseg,
  title={Gmmseg: Gaussian mixture based generative semantic segmentation models},
  author={Liang, Chen and Wang, Wenguan and Miao, Jiaxu and Yang, Yi},
  booktitle={NeurIPS},
  year={2022}
}

@article{xu2025jodi,
  title={Jodi: Unification of visual generation and understanding via joint modeling},
  author={Xu, Yifeng and He, Zhenliang and Kan, Meina and Shan, Shiguang and Chen, Xilin},
  journal={arXiv preprint arXiv:2505.19084},
  year={2025}
}

@article{yang2025vrmdiff,
  title={VRMDiff: Text-Guided Video Referring Matting Generation of Diffusion},
  author={Yang, Lehan and Song, Jincen and Wang, Tianlong and Qi, Daiqing and Shi, Weili and Liu, Yuheng and Li, Sheng},
  journal={arXiv preprint arXiv:2503.10678},
  year={2025}
}
}
\newpage
\appendix

\noindent
\textbf{Overview.} In the supplementary materials, we present more results and details:
\begin{itemize}
    \item \textbf{\ref{sec:appendix_details}.} More details on the method. 
    \item \textbf{\ref{sec:appendix_studies}.} More empirical studies and visual results.
    \item \textbf{\ref{sec:appendix_limit}.} Discussion of the broader impacts, safeguards and limitations.
    \item 
    \textbf{\ref{sec:appendix_compute}.} 
    Compute Reporting.
\end{itemize}

\section{Implementation Details}
\label{sec:appendix_details}

Our FlowSeg model is trained with 80K iterations for stage 1 and 40K iterations for stage 2 on the ADE20K dataset, and with 320K iterations for stage 1 and 160K iterations for stage 2 on the COCO-Stuff dataset. The maximum gradient norm is clipped to 1.0. To further validate the effectiveness of our method on a dataset with higher label cardinality, we additionally evaluate it on ADE20K-847. Unless otherwise specified, ADE20K refers to the standard 150-class setting (ADE20K-150).

During training, segmentation masks may include invalid regions, so we apply a binary mask $M$ to filter out these regions in both our method and the flow matching baselines, as defined in Eq.~\ref{eq:loss_w_mask}. In the main manuscript, we omit it for brevity in line with standard practices in the segmentation field.
\begin{equation}
\label{eq:loss_w_mask}
\mathcal{L}_{res} = \mathbb{E} \left[
\left\lVert
\left( v_\theta(x_t, t, I) - \text{sg}[\tilde{v}_t] \right) \odot M
\right\rVert^2
\right]
\end{equation}
During inference, the predicted mask is compared against the nearest semantic centroid, and the corresponding category is assigned accordingly.

We summarize the main hyperparameters in Tab.~\ref{tab:supp_hyperparam1} and Tab.~\ref{tab:supp_hyperparam2}. 
The main experiments use the full two-stage schedule, whereas ablations use only the first stage, trained for 20K iterations.

\begin{table}[h]
\centering
\small
\setlength{\tabcolsep}{6pt}
\caption{Setting in Stage 1.}
\label{tab:supp_hyperparam1}
\begin{tabular}{ll}
\toprule
\textbf{Hyperparameter} & \textbf{Value} \\
\midrule
Input resolution & $256 \times 256$ \\
Total batch size & 64 \\
\midrule
Backbone & PixNerd-XL-P16-C2I \\
Patch size & 16 \\
Hidden dimension & 1152 \\
Transformer blocks & 30 \\
Conditional blocks & 26 \\
Input / output channels & 3 / 3 \\
Pretraining & PixNerd-XL-P16-C2I checkpoint \\
\midrule
Optimizer & AdamW \\
Learning rate & $5\times10^{-5}$ \\
Adam betas & $(0.9, 0.999)$ \\
Weight decay & 0.01 \\
Gradient clipping & 1.0 \\
\bottomrule
\end{tabular}
\end{table}

\begin{table}[h]
\centering
\small
\setlength{\tabcolsep}{6pt}
\caption{Setting in Stage 2.}
\label{tab:supp_hyperparam2}
\begin{tabular}{ll}
\toprule
\textbf{Hyperparameter} & \textbf{Value} \\
\midrule
Input resolution & $512 \times 512$ \\
Total batch size & 16 \\
\midrule
Backbone & PixNerd-XL-P16-C2I \\
Patch size & 16 \\
Hidden dimension & 1152 \\
Transformer blocks & 30 \\
Conditional blocks & 26 \\
Input / output channels & 3 / 3 \\
Pretraining & Stage 1 checkpoint \\
\midrule
Optimizer & AdamW \\
Learning rate & $1\times10^{-5}$ \\
Adam betas & $(0.9, 0.999)$ \\
Weight decay & 0.01 \\
Gradient clipping & 1.0 \\
\bottomrule
\end{tabular}
\end{table}

\section{Empirical Results}
\label{sec:appendix_studies}

\subsection{Pixel Space Modeling}

\begin{figure}[ht]
	\centering
	\includegraphics[width=1.0\linewidth]{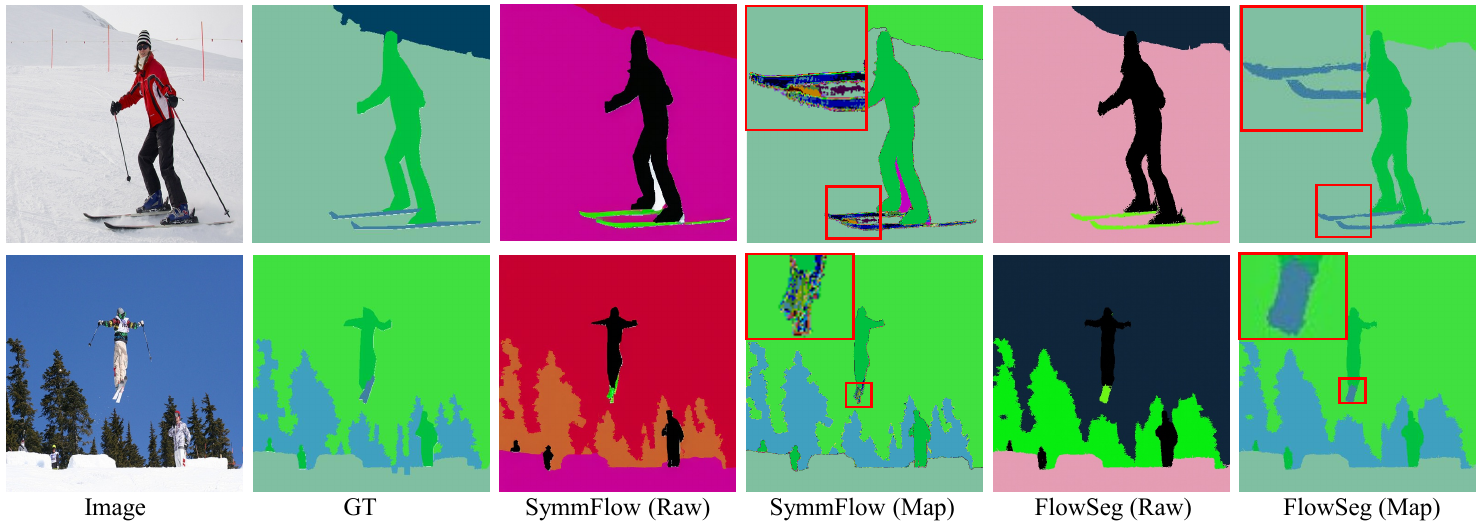}
	\caption{Visualization of the impact of VAE on the final prediction. \textbf{GT} denotes the ground truth. The \textbf{Raw} prediction is the model's direct, unprocessed output. To obtain the \textbf{Map} prediction, we compute the distance between each pixel and the candidate centroids, and assign each pixel to the category of its nearest centroid. In this comparison, SymmFlow uses latent-space modeling with a VAE, whereas our FlowSeg adopts end-to-end pixel-space modeling.}
    \label{fig:sup_vae_cmp}
\end{figure}

We employ end-to-end pixel-space modeling rather than traditional latent-space modeling with VAEs. The VAE framework constructs a distorted latent space for optimization, which diminishes pixel-level distinctions to the extent that they become indistinguishable. This characteristic makes VAEs inherently unsuitable for segmentation tasks.
As shown in Fig.~\ref{fig:sup_vae_cmp}, VAEs exacerbate this issue: although the predicted colors appear nearly identical, minor differences can lead to assignment to entirely different semantic categories, resulting in the \textit{scattered} effect highlighted in the red box.

\subsection{Convergence Analysis}

\begin{figure}[ht]
	\centering
	\includegraphics[width=0.65\linewidth]{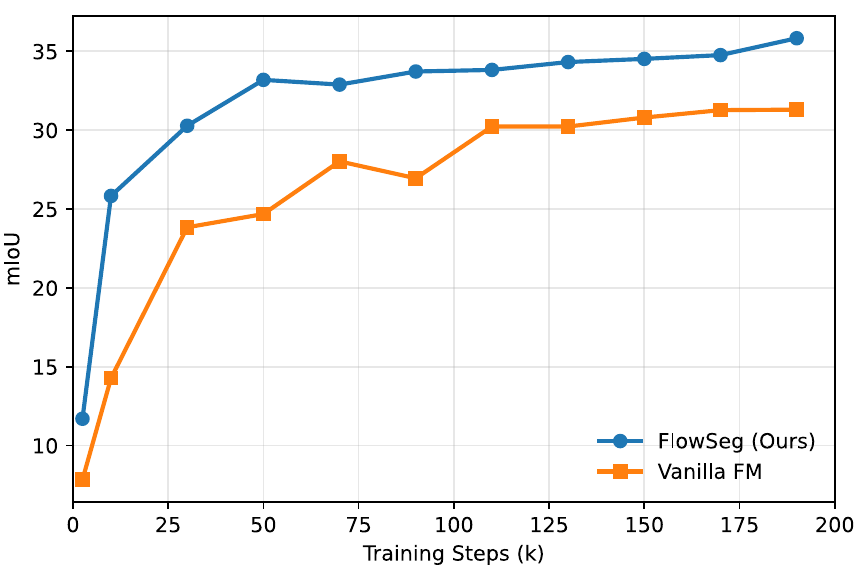}
	\caption{Comparison with vanilla FM on the ADE20K dataset under a long training schedule (200K iterations, batch size 64). The resolution is set to $256\times 256$.}
    \label{fig:sup_vsmse_long}
\end{figure}

To better highlight the differences between FlowSeg and vanilla flow matching, we extend the training schedule to 200K iterations with a batch size of 64, corresponding to over 600 epochs, ensuring convergence.
As shown in Fig.~\ref{fig:sup_vsmse_long}, FlowSeg consistently outperforms vanilla FM throughout the training process, demonstrating both faster convergence and superior performance.

\subsection{Scaling to Higher Cardinality}
We evaluate our method on ADE20K-847, a challenging semantic segmentation benchmark with 847 categories, substantially more than most commonly used datasets. We follow the two-stage training procedure described in the main manuscript. The results are reported in Tab.~\ref{tab:ade847}.

\begin{table}[t]
\centering
\small
\setlength{\tabcolsep}{8pt}
\caption{Comparison on ADE20K-847.}
\label{tab:ade847}
\begin{tabular}{llc}
\toprule
Model & Type & mIoU \\
\midrule
SimSeg & Discriminative Specialists & 10.43 \\
PromptDiffusion & Diffusion Models & 0.99 \\
FlowSeg (Ours) & Diffusion Models & \textbf{10.77} \\
\bottomrule
\end{tabular}
\end{table}

The results demonstrate that FlowSeg remains effective under extreme class cardinality, achieving performance comparable to the specialized discriminative model SimSeg, while substantially outperforming the diffusion-based baseline PromptDiffusion, which struggles in this setting.

\subsection{Sensitivity to Temperature}
\label{sec:supp_temp_coef}

We first evaluate the temperature parameter $\tau$ within the softmax assignment. As shown in Tab.~\ref{tab:supp_temp}, the optimization dynamics are sensitive to this scaling. Low temperature yields an overly sharp assignment, destabilizing the gradient and severely bottlenecking performance. Conversely, excessive temperatures flatten the distance-aware penalty, diluting the discriminative signal. The sweet spot of $\tau$ is around $0.5$ to $1$, indicating that the reshape field relies on moderate sharpness to deliver gradients that are both semantically informative and numerically stable.

\begin{table}[h]
\centering
\small
\setlength{\tabcolsep}{10pt}
\caption{\textbf{Temperature ablation.} We vary the temperature $\tau$ in the distance-to-probability transformation while keeping other settings constant. Moderate sharpness yields the best performance.}
\label{tab:supp_temp}
\begin{tabular}{lccccc}
\toprule
Dataset & 0.25 & 0.50 & 1.00 & 2.00 & 5.00 \\
\midrule
ADE-150 & 20.17 & 27.29 & \textbf{29.75} & 23.36 & 19.73 \\
ADE-847 & 2.75 & \textbf{4.59} & 4.53 & 3.15 & 3.02 \\
\bottomrule
\end{tabular}
\end{table}

\subsection{Centroid Assignment}

We conduct an ablation study on centroid assignment. Specifically, we treat the previously ignored category as an additional valid class, resulting in a total of $ K+1$ centroids. During inference, although the model outputs raw predictions for all centroids, the nearest label assignment is performed only among the $K$ valid centroids. As shown in Tab.~\ref{tab:sup_kplus}, including an additional centroid has a limited impact on performance (Rows $1^{st}$ and $3^{rd}$). Instead, performance is more sensitive to mask filtering (Eq.~\ref{eq:loss_w_mask}), which is consistent with our conclusions in the main manuscript.

We conduct experiments on the ADE-150 dataset, considering two training recipes: vanilla flow matching and FlowSeg, and two initialization methods: Kronecker-style and PCA-style. The latter incorporates semantic information. Specifically, for each category in the ADE-150 dataset, we used the prompt "a photo of \{CLASS NAME\}" to extract text embeddings via CLIP ViT-B, then applied PCA to reduce the dimensionality to three and normalized the results to [-1, 1]. The results are shown in Tab.~\ref{tab:abla_centroid_init}.

\begin{table}[ht]
\centering
\caption{Ablations on centroid assignment. Additional-Centroid indicates assigning a specific centroid for the ignored category.}
\label{tab:sup_kplus}
\begin{tabular}{ccc}
\toprule[0.1em]
Additional-Centroid & Mask & mIoU \\
\midrule[0.1em]
- & \checkmark & 29.75 \\
\checkmark & - & 26.80 \\
\checkmark & \checkmark & 29.21 \\
\bottomrule[0.1em]
\end{tabular}
\end{table}

\begin{table}[htbp]
\centering
\small
\caption{Effect of centroid initialization on vanilla FM and FlowSeg.}
\label{tab:abla_centroid_init}
\begin{tabular}{llcc}
\toprule
Method & Init. & mIoU & Min. NN Dist. \\
\midrule
Vanilla FM & PCA & 24.16 & 0.02 \\
Vanilla FM & Kronecker & 21.14 & 0.15 \\
FlowSeg & PCA & 28.42 & 0.02 \\
FlowSeg & Kronecker & 29.75 & 0.15 \\
\bottomrule
\end{tabular}
\end{table}

\subsection{Learning Rate Compensation}

\begin{figure}[ht]
	\centering
	\includegraphics[width=0.7\linewidth]{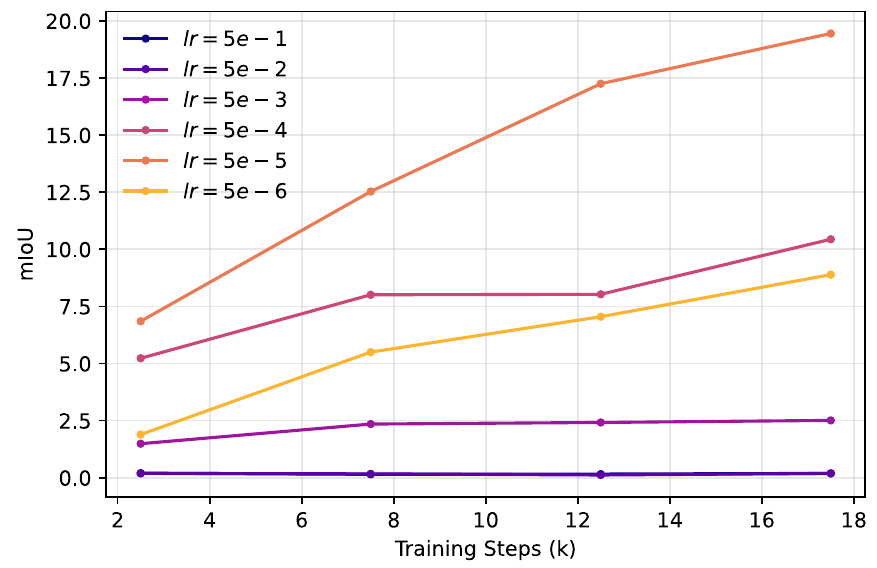}
	\caption{Vanilla FM with different learning rates. $5e-5$ is the default setting.}
    \label{fig:sup_lr}
\end{figure}

We observe that the difference in convergence, particularly the gradient vanishing phenomenon, between our FlowSeg and vanilla flow matching stems from the underlying optimization dynamics and cannot be mitigated simply by increasing the learning rate. As shown in Fig.~\ref{fig:sup_lr}, naively raising the learning rate results in significant performance degradation. This is because an excessively high learning rate causes regions far from the centroid to experience extremely large updates, leading to severe optimization oscillations and preventing convergence to the optimal solution. On the other hand, a learning rate that is too low also results in suboptimal performance due to insufficient optimization.

\subsection{Generative Property Analysis}
\begin{figure}[ht]
	\centering
	\includegraphics[width=0.8\linewidth]{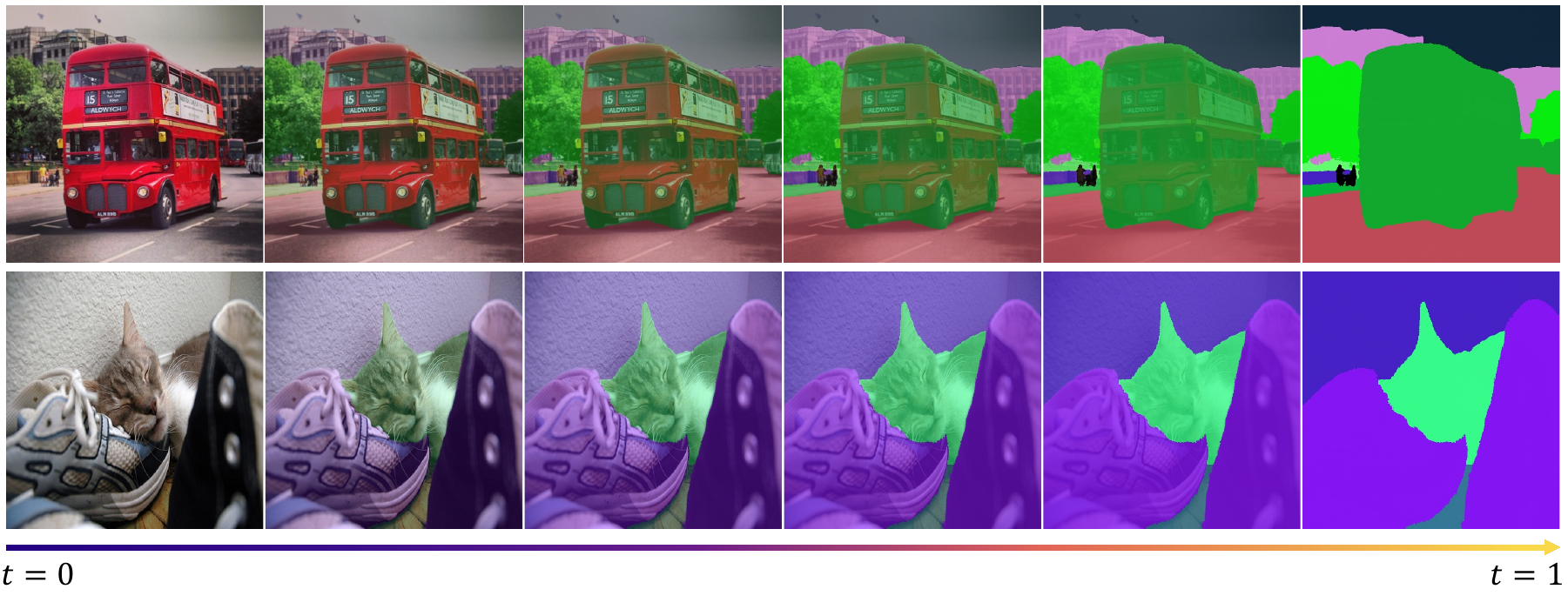}
	\caption{Visualization of intermediate states during sampling.}
    \label{fig:abla_samplevis}
\end{figure}

FlowSeg inherits the intrinsic characteristics of generative models. Built upon the ODE framework of flow matching, we employ an Euler solver to evaluate performance across various sampling steps (Tab.~\ref{tab:abla_infer_Step}). On the ADE20K dataset, FlowSeg achieves peak performance at 10 steps; however, efficiency-accuracy trade-offs arise, as too few steps lead to discretization errors, while excessive steps may accumulate approximation errors. As visualized in Fig.~\ref{fig:abla_samplevis}, the intermediate sampling states exhibit smooth evolution trajectories between the image and mask distributions, confirming that our model preserves the continuous mapping properties of flow-based generation.

\begin{table}[!ht]
\centering
\caption{Sampling with different steps on the ADE20K dataset. Euler sampler is adopted.}
\label{tab:abla_infer_Step}
\begin{tabular}{lcccccccc}
\toprule[0.1em]
Step & 1 & 2 & 5 & 10 & 20 & 30 & 50 & 100 \\ \hline
mIoU & 43.47 & 45.85 & 46.82 & \textbf{47.1} & 46.78 & 46.59 & 46.13 & 45.78 \\
\bottomrule[0.1em]
\end{tabular}
\end{table}

\section{Discussion}
\label{sec:appendix_limit}

\subsection{Broader Impacts} 

This work explores and analyzes the limitations of traditional diffusion-based segmentation methods and proposes a novel vector field reshaping approach that significantly improves the model's discriminative ability while preserving generation capability. We believe this method holds promise for further bridging the gap between generation and understanding, and ultimately promoting a unified solution. 

\subsection{Safeguards}

Given that the primary task of this work is perception and that the generated data are mostly segmentation masks, it does not produce harmful content compared to standard generative models. Furthermore, the datasets used are publicly available and have been extensively validated by the community.

\subsection{Limitations}

Given that this work primarily explores the feasibility of the concept, we do not focus on architectural modifications or the scaling laws that have been proven effective in engineering, and some segmentation failures exist. However, we believe this method has not yet reached its limits; a better architectural design could achieve even more powerful results, which we leave for future work.

\section{Compute Reporting}
\label{sec:appendix_compute}

All experiments are conducted on a single-node cluster equipped with Intel(R) Xeon(R) Gold 6248 CPUs @ 2.50GHz, 8 NVIDIA A100 GPUs, and 1.5 TB of system memory.

\end{document}